\documentclass{article}
\usepackage[final,main]{neurips2026new8}
\usepackage[utf8]{inputenc}
\usepackage[T1]{fontenc}
\usepackage{hyperref}
\usepackage{url}
\usepackage{booktabs}
\usepackage{amsfonts}
\usepackage{nicefrac}
\usepackage{microtype}
\usepackage{xcolor}
\usepackage{graphicx}
\usepackage{tikz}
\usetikzlibrary{positioning, arrows.meta}

\title{IPO Finance Agent: Benchmark of LLM Financial Analysts Beyond Finance Agent v2, with Automated Rubric Generation, on the SpaceX (SPCX) IPO}

\author{%
  Mostapha Benhenda \\
  \texttt{mostaphabenhenda@gmail.com} \\
}

\begin{document}

\makeatletter
\renewcommand{\@noticestring}{}
\makeatother

\maketitle

\begin{abstract}
Finance Agent v2 (by Vals AI) has emerged as the reference benchmark for evaluating both Anthropic Claude and OpenAI ChatGPT frontier language models on financial tasks. However, it narrowly deals with periodic reporting from publicly traded companies (SEC 10-K and 10-Q filings), and its agentic harness relies on naive, unenriched chunk retrieval. Neither the task design nor the retrieval approach addresses the distinct challenges of IPO due diligence. SEC S-1 filings combine historical financial statements, governance structures, pro forma and common-control accounting treatments, capital-formation narratives, and underwriting-sensitive risk disclosures within substantially longer documents than typical periodic filings.

That is why we introduce \textbf{IPO Finance Agent}, which extends the Finance Agent v2 framework along two directions: task domain and retrieval architecture. During our experiments, the original Finance Agent v2 harness basically failed to deliver any output related to the SpaceX S-1 filing, due to document length. We therefore had to improve the agentic harness with contextual retrieval, a more realistic and industry-standard approach for long documents. We also built a dataset of 1{,}000 IPO-diligence questions, and publicly release 70 questions on the SpaceX (SPCX) S-1 filing to support reproducibility, while the remainder are held private to guard against benchmark contamination.

In addition, we introduce an evaluator-optimizer pipeline to automatically generate evaluation rubrics for the benchmark: candidate facts are extracted from model answers, consolidated into draft criteria, then automatically audited for omissions, hallucinations, mistiered items, and redundancy, with LLM feedback driving iterative repair, targeted enrichment, and deduplication. Human experts only review final rubrics before deployment.

Results show that the best-performing evaluated model, Zhipu GLM-5.2, reaches 79.8\% accuracy, and the most cost-efficient model on the resulting Pareto frontier, Xiaomi MiMo-2.5 Pro, reaches slightly lower accuracy (77.2\%) at 0.05 USD per query, while exceeding the current Finance Agent v2 leaderboard ceiling, Google Gemini 3.5 Flash at 57.9\% for 2.51 USD per query, and undercutting even FABv2's cheapest entry (MiniMax M3: 48.3\% at 0.32 USD) on cost-efficiency.
Code and data are released on GitHub \url{https://github.com/benstaf/ipoagent}
\end{abstract}

\section{Introduction}
\label{sec:intro}

Large language models are increasingly deployed as financial research assistants, tasked with extracting, reconciling, and reasoning over disclosures that previously required hours of analyst time. Recent agent evaluation research has shown that benchmark outcomes often depend not only on model capability but also on retrieval architecture, tool interfaces, environment design, and evaluation methodology \cite{liang2022helm,liu2023agentbench,yao2024taubench,jimenez2024swebench}. This observation is particularly relevant in financial analysis, where the evidence required to answer a question may be dispersed across lengthy filings and external data sources. The Finance Agent benchmark (by Vals AI) has become a widely used reference for measuring this capability, used by both Anthropic and OpenAI to report frontier model performance on SEC-filing-grounded tasks such as ratio extraction, comparables construction, and earnings reconciliation \cite{financeagent,financeagentv2}. Its second iteration, Finance Agent v2, sharpened this evaluation considerably: a nine-category taxonomy reorganized around real analyst workflows, dealbreaker-gated partial-credit grading, multi-run aggregation, and a private held-out test split. Even so, the entire benchmark---like the periodic-reporting workflows it models---is built around mature, already-public companies filing routine 10-K and 10-Q disclosures.

Initial public offering (IPO) due diligence is a different task. The accounting and finance literature has long documented that IPO prospectuses differ materially from periodic reports because of extensive use of pro forma disclosures, related-party transactions, governance disclosures, and underwriter-mediated risk-factor construction \cite{ritter1991,lowry2017ipo,ljungqvist2007ipo}. An S-1 SEC registration statement for IPO is not an update to a known financial profile; it is often the first complete disclosure of a company's economics, and it must be read without the benefit of prior-year guidance, established Wall Street consensus, or a multi-year disclosure history to anchor interpretation. S-1 filings layer historical financial statements together with governance and control structures (multiple share classes, founder voting control, board composition), pro forma and common-control recast accounting (frequently required when a private company has undergone reorganizations or affiliate transactions before filing), forward-looking capital formation narratives, and underwriting-sensitive risk factors written for a transaction that has not yet priced. Reasoning correctly about any one of these elements typically requires reasoning about several simultaneously---for example, interpreting a recast income statement requires understanding both the accounting mechanics of common-control combinations and the governance narrative explaining why the recast was necessary in the first place. These documents are also substantially longer than a typical 10-K or 10-Q, which compounds the retrieval problem: the evidence needed to answer a single question is frequently scattered across distant sections of the filing---a risk factor, an MD\&A paragraph, and a footnote to the pro forma financial statements---in a way that naive fixed-window chunking is poorly suited to surface. During our experiments, we found that the original Finance Agent v2 harness did not even complete evaluation on the Space X long S-1 filing. This motivated a change in retrieval architecture before model comparison experiments were conducted.

This combination of domain shift and document length shift means that strong performance on Finance Agent v2's periodic reporting tasks does not necessarily transfer to IPO diligence, and existing benchmarks offer no direct way to measure whether it does. We introduce \textbf{IPO Finance Agent} to fill this gap, using the SpaceX (SPCX) S-1 as a case study---a filing of particular interest given its huge impact, unusual segment structure (Starlink, the core launch business, and adjacent ventures including xAI exposure) and its reliance on secondary-market pricing data (Forge Global, Nasdaq Private Market) in place of a public trading history.

Our contributions are threefold:

\begin{enumerate}
\item \textbf{A benchmark of IPO diligence questions} spanning seven domains---segment economics and operating performance, KPI quality and monetization, governance and control structures, accounting and common-control recast mechanics, capital intensity and funding, execution and program risk, and valuation and underwriting analysis---each labeled by the professional workflow it reflects (investment banking and equity capital markets, public market investing, venture capital and growth investing, credit analysis, securities counsel, and accounting/transaction advisory). Of 1{,}000 total questions, 70 covering the SpaceX S-1 filing are released publicly; the remainder are held privately to limit benchmark contamination, like Finance Agent v2 did, consistent with broader concerns about benchmark contamination in LLM evaluation \cite{golchin2024benchmark}.
\item \textbf{A contextual-retrieval harness} that replaces the naive, unenriched chunk retrieval used in Finance Agent v2's public implementation. We add context-enriched embeddings, addressing the specific failure mode of long, cross-referential registration statements, where relevant evidence is distant from the question's surface phrasing.
\item \textbf{An automated, evaluator-optimizer pipeline for rubric generation} that produces and audits grading rubrics without manual annotation: candidate facts are extracted from model answers, then a consolidation stage merges and normalizes them into draft grading criteria, and an evaluator stage iteratively critiques the draft---flagging structural defects, coverage gaps, and redundant entries---routing structural defects to a scope-limited repair step and coverage gaps to a targeted re-extraction (enrichment) step, with a deduplication pass and a final human expert review pass before deployment.
\end{enumerate}

The remainder of this paper is organized as follows. Section~\ref{sec:related} situates IPO Finance Agent relative to Finance Agent v2 and the broader landscape of financial NLP benchmarks and agentic evaluation methodology. Section~\ref{sec:dataset} describes the SpaceX S-1 case study and the seven domain question taxonomy. Section~\ref{sec:method} details the contextual-retrieval harness and the rubric generation pipeline. Section~\ref{sec:results} reports model performance and cost-accuracy tradeoffs. Section~\ref{sec:discussion} examines what the performance gap relative to Finance Agent v2 suggests about harness design versus task difficulty, and Section~\ref{sec:conclusion} concludes.

\section{Related Work}
\label{sec:related}

\paragraph{Financial QA and financial LLM benchmarks.} Prior to agentic evaluation, financial NLP benchmarks such as FinQA \cite{chen2021finqa}, TAT-QA \cite{zhu2021tatqa}, and ConvFinQA \cite{chen2022convfinqa} established the core challenge of numerical reasoning over financial documents---extracting figures from tables and text and chaining them through arithmetic operations to answer a question. These benchmarks were largely single-turn and single-document, testing extraction and computation rather than multi-step research. FinanceBench \cite{financebench} extended this to longer 10-K-style documents but retained a primarily extractive framing. More recently, FinBen introduced a holistic financial benchmark spanning 42 datasets and 24 financial tasks, including retrieval-augmented and agent-based evaluation settings \cite{finben}, and MultiFinBen extended financial evaluation into multilingual and multimodal settings, evaluating LLMs across text, vision, and audio modalities on cross-lingual financial reasoning tasks \cite{multifinben}. IPO Finance Agent differs from this entire family of benchmarks in its focus on IPO due diligence and S-1 registration statements, a disclosure regime characterized by governance analysis, common-control accounting, capital-formation narratives, and underwriting-sensitive disclosures that none of these benchmarks directly target.

\paragraph{Finance Agent (by Vals AI) benchmarks.} Finance Agent (by Vals AI) marked a shift toward \emph{agentic} evaluation: rather than presenting a model with a pre-selected document excerpt, it requires the model to search, retrieve, and synthesize across full filings and external data sources (price history, EDGAR) using a tool-use harness, with grading against expert-authored rubrics rather than exact-match answers \cite{financeagent}. Finance Agent v2 deepened this further with a nine-category taxonomy---General Qualitative Analysis, General Quantitative Analysis, Market Analysis, Comparables, Precedents, Adjustments, Earnings Analysis, Disclosure Analysis, and Financial Modeling---reorganized around real analyst workflows rather than retrieval difficulty tiers, dealbreaker-gated partial-credit grading, and a held-out private test split to limit contamination \cite{financeagentv2}. Closely related recent benchmarks target adjacent agentic-finance capabilities: FinRetrieval evaluates an agent's ability to retrieve specific numeric values from structured financial databases and finds that tool availability, more than model sophistication, drives performance \cite{finretrieval}, while WorkstreamBench evaluates agents on end-to-end spreadsheet construction for financial modeling and scenario analysis \cite{workstreambench}. IPO Finance Agent adopts the agentic, rubric-graded evaluation philosophy of Finance Agent v2 but applies it to a document type and disclosure regime---first-time registration statements---that none of these benchmarks cover.

\paragraph{Industry adoption of Finance Agent (by Vals AI)} Finance Agent has evolved beyond an academic benchmark into an evaluation that frontier model developers themselves report. OpenAI's GPT-5.5 launch reports a Finance Agent score of 60.0\% \cite{openai_gpt55}. Anthropic has likewise reported Finance Agent results across successive financial services product announcements and model launches: Claude for Financial Services \cite{anthropic_finance_2025}; Advancing Claude for Financial Services, which reported Claude Sonnet 4.5 topping the benchmark at 55.3\% accuracy \cite{anthropic_finance_2025b}; and the Claude Opus 4.8 launch, which reports Opus 4.8 leading Finance Agent v2 among frontier-class models at 53.9\% \cite{anthropic_opus48}. Moonshot AI has likewise publicized Finance Agent v2 results, tweeting that Kimi K2.6 topped the benchmark among open-weight models \cite{moonshot_financeagentv2}. This adoption by multiple frontier model providers means that Finance Agent has emerged as a de facto reference evaluation for agentic financial research tasks, and motivates extending its evaluation philosophy to disclosure regimes, such as IPO registration statements, which it does not cover yet.

\paragraph{Venture Capital and startup evaluation.} Beyond public market finance, a parallel line of benchmarks targets earlier stages of the company lifecycle. VCBench evaluates LLMs on founder-success prediction using a large set of anonymized founder profiles, framing venture investing itself as a forecasting task \cite{vcbench}. YC Bench similarly evaluates the prediction of startup outperformance within Y Combinator batches using publicly observable pre-demo-day signals \cite{ycbench}. Together, these benchmarks span seed-stage investing and founder selection, whereas IPO Finance Agent focuses on the disclosure-intensive transition from private to public markets---a complementary stage in the same continuum of company financing.

\paragraph{Agent benchmarks and holistic evaluation.} IPO Finance Agent is also related to a broader line of work evaluating language models as agents rather than static question-answering systems. HELM introduced the notion of holistic evaluation across diverse scenarios and metrics, emphasizing that benchmark design itself can substantially influence measured model capabilities \cite{liang2022helm}. BIG-Bench similarly demonstrated the value of broad, capability-oriented benchmark suites spanning hundreds of heterogeneous tasks contributed by a large research community \cite{srivastava2023bigbench}. AgentBench formalized evaluation of language models acting in interactive environments, demonstrating that agent performance depends on tool use, planning, and long-horizon reasoning in addition to language generation quality \cite{liu2023agentbench}. More recent benchmarks such as $\tau$-bench evaluate tool-agent-user interaction under realistic operational constraints \cite{yao2024taubench}, while SWE-bench evaluates end-to-end software engineering problem solving from real GitHub issues and has become a reference benchmark for coding agents \cite{jimenez2024swebench}, and GAIA evaluates general purpose assistant capabilities across multi step, tool-dependent tasks \cite{mialon2024gaia}. IPO Finance Agent occupies a complementary position within this literature: rather than evaluating general purpose agency, customer service workflows, or software engineering, it evaluates professional financial due diligence over long-form SEC registration statements, combining challenges from financial document understanding with challenges from agentic retrieval, evidence synthesis, and workflow-grounded reasoning.

\paragraph{IPO and S-1-specific analysis.} Outside of LLM benchmarking, the accounting and finance literature has long treated IPO prospectus analysis as a distinct discipline from periodic financial statement analysis. IPO registration statements serve as a central mechanism for reducing information asymmetry between issuers and investors, while simultaneously acting as vehicles for governance disclosure, risk communication, and valuation framing \cite{ritter1991,ritter2002,ljungqvist2007ipo,lowry2017ipo}. The prevalence of pro forma reporting, common-control transactions, related-party disclosures, and underwriter-mediated risk-factor drafting further distinguishes S-1 analysis from periodic reporting workflows. To our knowledge, no existing LLM benchmark targets this disclosure regime directly; IPO Finance Agent being the first agentic benchmark constructed specifically around S-1 diligence tasks.

\paragraph{Retrieval over long financial documents.} Naive fixed-size chunking for retrieval-augmented generation \cite{lewis2020rag} and dense retrieval pipelines built on passage-level embeddings \cite{karpukhin2020dpr} can degrade when relevant context spans chunk boundaries or when a chunk's relevance depends on information stated elsewhere in the document---a failure mode that becomes more severe as document length grows. Anthropic's contextual retrieval approach addresses this by prepending chunk-specific, document-level context (generated by an LLM) to each chunk before embedding and indexing, which has been shown to meaningfully reduce retrieval failure rates on long, heterogeneous documents \cite{anthropic_contextual_retrieval}. This can be viewed as complementary to late-interaction retrieval architectures such as ColBERT \cite{khattab2020colbert}, which similarly aim to preserve fine-grained retrieval fidelity on long or structurally complex documents. S-1 filings are a natural stress test for this technique: a single question about, say, segment-level capital intensity may require connecting a risk-factor paragraph, an MD\&A discussion, and a footnote to the financial statements that share no lexical overlap with the question itself. We adopt contextual retrieval as the core architectural change distinguishing our harness from Finance Agent v2's public implementation, which performs retrieval over unenriched chunks.

\paragraph{Automated rubric and evaluation construction.} Constructing grading rubrics for open-ended, multi-step research tasks is itself an open problem: Finance Agent v2's rubrics are human expert-authored and reviewed by a panel of financial professionals, a process that is accurate but does not scale easily to new document types or larger question sets. We instead adopt the evaluator-optimizer pattern described in Anthropic's agentic-workflow taxonomy \cite{anthropic_agents}, in which one model call generates a candidate artifact and a second model call evaluates it against explicit criteria, feeding that evaluation back as the input to a further generation step, iterated until the artifact converges or a stopping criterion is reached. This approach is closely related to recent work on LLM-based evaluation, self-refinement, constitutional feedback loops, and LLM-as-a-judge methodologies \cite{bai2022constitutional,madaan2024selfrefine,zheng2023,liu2023geval,zhu2024judgelm}. We apply this pattern to rubric construction itself: candidate facts are extracted from an ensemble of independently-generated model answers and consolidated into a draft rubric, which is then iteratively evaluated and routed to either scope-limited structural repair or targeted re-extraction (enrichment), depending on whether the identified defect is structural or evidentiary---before a final deduplication pass and human expert review. This positions our contribution not only as a new benchmark but as a methodology for constructing graded benchmarks from complex source documents with substantially reduced manual annotation effort.

\section{Dataset Construction}
\label{sec:dataset}

\subsection{Case Study Selection: The SpaceX S-1 SEC filing}

We use the SpaceX S-1 SEC filing \cite{spaceexplorationtechnologies2026s1} as our flagship case study for the publicly released portion of the benchmark for three reasons, beyond the wide publicity around this IPO. First, its segment structure is unusually complex for a first-time filer: launch services, Starlink, and adjacent exposure (including the company's stake-related disclosures touching xAI and X) sit alongside one another in ways that require cross-segment reasoning rather than single-segment extraction. Second, the filing contains common-control recast financial statements---historical results restated to reflect entities under common control prior to the offering---which is a recurring and underspecified challenge in IPO accounting that has no analogue in standard 10-K/10-Q analysis. Third, because SpaceX equity has traded for years on secondary markets prior to filing, valuation questions cannot rely on a public trading history the way a typical periodic-filing question can; instead they require triangulating Forge Global and Nasdaq Private Market secondary pricing data against primary-filing disclosures, peer comparables (Rocket Lab, Eutelsat, Viasat, Amazon Kuiper, Meta), and the S-1's own narrative framing. Together these properties make the SpaceX S-1 a demanding but representative stress test for the broader IPO-diligence task class. The private portion of the dataset draws on a further set of recent S-1 filings spanning multiple sectors, to avoid over-indexing the full 1{,}000-question benchmark on a single company's idiosyncrasies.

\subsection{Question Taxonomy: Seven IPO Diligence Domains}

Questions are organized into seven domains, chosen to span the analytical surface area of an S-1 rather than to mirror the retrieval difficulty framing common in periodic filing benchmarks, including Finance Agent v2:

\begin{itemize}
\item \textbf{Segment economics and operating performance}---unit economics, margin structure, and capital intensity by segment, including cross-segment cost allocation where disclosed.
\item \textbf{KPI quality and monetization}---scrutiny of company-defined operating metrics (e.g., active subscriber counts, backlog, launch cadence) for consistency, comparability across periods, and resistance to favorable framing.
\item \textbf{Governance and control structures}---share class structure, founder voting control, board composition, and related-party governance provisions.
\item \textbf{Accounting and common-control recast mechanics}---the accounting judgment underlying pro forma adjustments and common-control combinations, and their effect on reported historical performance.
\item \textbf{Capital intensity and funding}---forward capital requirements, identified funding sources, and their quantified or implied financial impact on the offering's use-of-proceeds narrative.
\item \textbf{Execution and program risk}---supply-chain risk factors, schedule risk in major programs, and the operational dependencies that could materially affect forward performance.
\item \textbf{Valuation and underwriting analysis}---comparable-company and secondary-market-based valuation reasoning, and interpretation of underwriting-sensitive risk disclosures.
\end{itemize}

\subsection{Professional Workflow Labeling}

Each question is additionally labeled with the professional workflow most likely to motivate it---investment banking and equity capital markets, public-market investing, venture capital and growth investing, credit analysis, securities counsel, or accounting and transaction advisory---reflecting the fact that the same disclosure can be read for different purposes (e.g., a recast income statement matters to an ECM banker assessing offering-readiness and to a credit analyst assessing pro forma leverage capacity for entirely different reasons). This labeling is independent of the seven-domain taxonomy and supports slicing benchmark results by audience in addition to by analytical domain.

\subsection{Public/Private Split and Contamination Control}
\label{sec:contamination}

Of the full 1{,}000-question dataset, 70 questions cover the SpaceX S-1 specifically and are released publicly to support reproducibility and independent verification of the benchmark's grading methodology. The remaining questions are held private, following the split convention established by Finance Agent v2's public/private-validation/test structure, to limit the risk that future model training runs could memorize benchmark answers rather than genuinely reason over the source filings, a challenge that has become increasingly prominent in modern LLM benchmark design \cite{golchin2024benchmark}.

\section{Methodology}
\label{sec:method}

\subsection{Agentic Harness and Contextual Retrieval}

Agents are evaluated within a tool-use harness equipped with web search and EDGAR access, following Finance Agent v2. The SEC's EDGAR system has become the primary infrastructure through which public-company disclosures are disseminated and analyzed by investors, researchers, and financial-information systems \cite{edgar}. The principal architectural change concerns retrieval. The public Finance Agent v2 implementation performs retrieval over fixed-size, unenriched chunks of filing text. In our experiments, this configuration failed to apply to the SpaceX filing. We therefore replaced it with contextual retrieval \cite{anthropic_contextual_retrieval}, an extension of dense retrieval architectures \cite{karpukhin2020dpr} in which each chunk is augmented with a short, document-aware description before embedding and indexing. This industry best practice is particularly suitable for the structure of S-1 filings: a chunk drawn from a footnote to the pro forma financial statements is often uninterpretable in isolation, but becomes retrievable against a question about, say, common-control recast mechanics once it carries context identifying it as part of that recast disclosure. Contextual retrieval itself follows established methodology; the failure mode it addresses here---naive chunking becoming unreliable once filing length exceeds what it can index---is one that periodic 10-K/10-Q filings rarely reach but that S-1 registration statements, with their combined historical, pro forma, and narrative sections, frequently do.

\subsection{Automated Evaluation Pipeline: Overview}

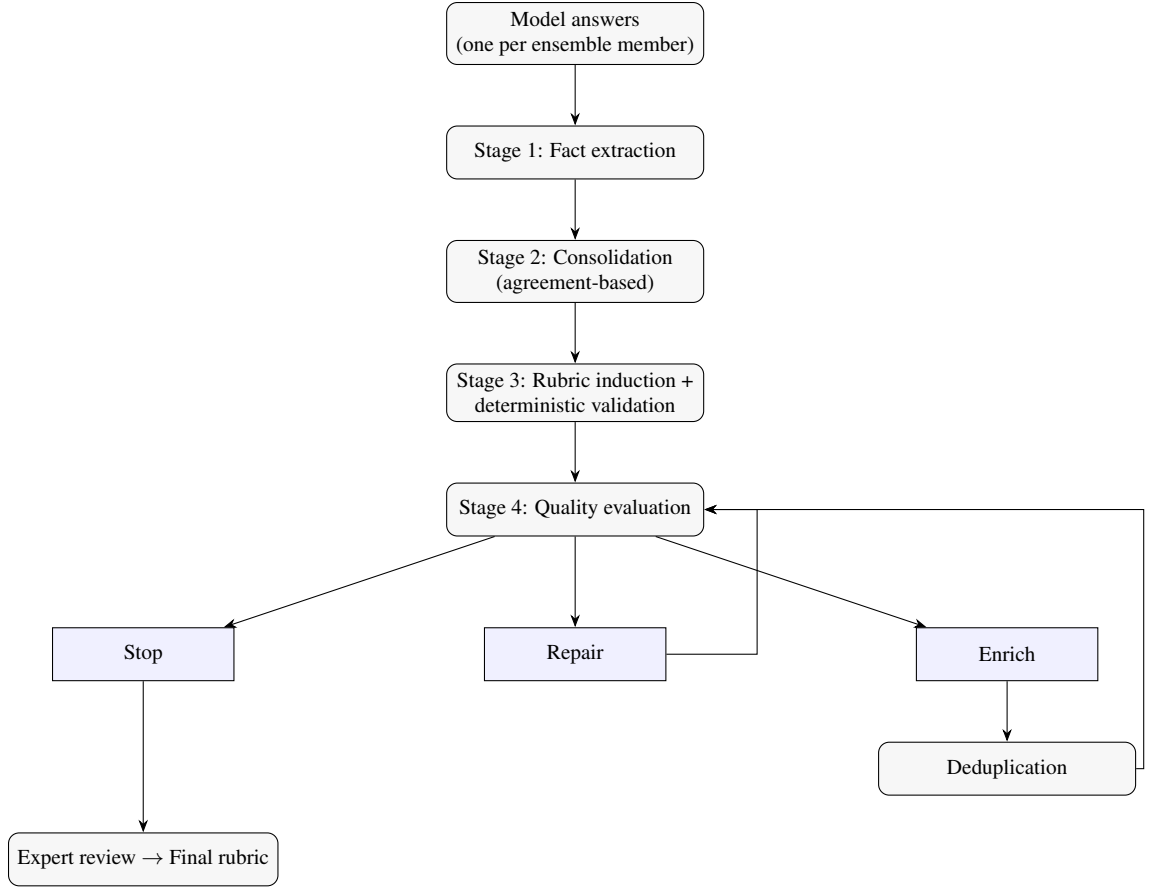
\begin{figure}[h]
\centering
\begin{tikzpicture}[
  node distance=8mm and 10mm,
  every node/.style={font=\footnotesize},
  box/.style={draw, rounded corners, align=center, minimum height=7mm, minimum width=34mm, fill=gray!6},
  decision/.style={draw, align=center, minimum height=7mm, minimum width=24mm, fill=blue!6},
  >=Stealth
]
\node[box] (answers) {Model answers\\(one per ensemble member)};
\node[box, below=of answers] (s1) {Stage 1: Fact extraction};
\node[box, below=of s1] (s2) {Stage 2: Consolidation\\(agreement-based)};
\node[box, below=of s2] (s3) {Stage 3: Rubric induction +\\deterministic validation};
\node[box, below=of s3] (s4) {Stage 4: Quality evaluation};
\node[decision, below left=12mm and 28mm of s4] (stop) {Stop};
\node[decision, below=12mm of s4] (repair) {Repair};
\node[decision, below right=12mm and 28mm of s4] (enrich) {Enrich};
\node[box, below=of enrich] (dedup) {Deduplication};
\node[box, below=20mm of stop] (expert) {Expert review $\rightarrow$ Final rubric};
\draw[->] (answers) -- (s1);
\draw[->] (s1) -- (s2);
\draw[->] (s2) -- (s3);
\draw[->] (s3) -- (s4);
\draw[->] (s4) -- (stop);
\draw[->] (s4) -- (repair);
\draw[->] (s4) -- (enrich);
\draw[->] (enrich) -- (dedup);
\draw[->] (repair.east) -- ++(12mm,0) |- (s4.east);
\draw[->] (dedup.east) -- ++(1mm,0) |- (s4.east);
\draw[->] (stop) -- (expert);
\end{tikzpicture}
\caption{The rubric-construction pipeline. After Stage 4 quality evaluation, the rubric is routed to \emph{stop} (expert review), \emph{repair} (structural fixes, rechecked directly), or \emph{enrich} (new evidence, followed by deduplication before rechecking). Repair and enrichment both loop back to Stage 4 for re-evaluation.}
\label{fig:pipeline}
\end{figure}

Grading rubrics for each question are produced by a multi-stage pipeline, an iterative evaluator-optimizer loop, following recent agentic workflow patterns for iterative generation and critique \cite{anthropic_agents,madaan2024selfrefine}: after an initial rubric is induced, an evaluator stage scores it against four quality dimensions and returns one of three recommendations---\emph{stop} (the rubric meets threshold and proceeds directly to expert review), \emph{repair} (structural defects fixable without new evidence, rechecked directly against the evaluator), or \emph{enrich} (coverage gaps requiring additional evidence from the original model answers, followed by a deduplication pass before being rechecked). Repair and enrichment are distinct branches with different inputs, different constraints, and different paths back to re-evaluation, detailed below and illustrated in Figure~\ref{fig:pipeline}.

\textbf{Stage 1---Fact extraction.} We take the answers given by GLM-5.1, GLM-5.2, Qwen 3.7 Max, Kimi K2.6, and MiMo-2.5 Pro. NVIDIA Nemotron 3 Ultra was excluded from the ensemble after an excessive number of failures, which compromised the reliability of its baseline answers for fact extraction. We then take an extraction model, GLM-5.1, to process each of these answers, pulling out atomic facts under a prompt that requires one verifiable claim per fact, preserves hedging language as written, and explicitly permits derived calculations (e.g., computing a margin from disclosed revenue and cost lines) rather than restricting extraction to verbatim-stated figures.

\textbf{Stage 2---Fact consolidation and normalization.} Extracted facts are merged across the ensemble's answers through a hybrid deterministic-and-incremental-LLM consolidation step: quantitative facts are clustered and canonicalized using deterministic grouping where possible, while qualitative facts are merged through incremental LLM-based consolidation. Facts are tagged with the set of models whose answers agreed on them, and only facts meeting a minimum-agreement threshold between models from different labs (agreements between GLM-5.1 and GLM-5.2 don't count), taken equal to 2 in our experiments are carried forward as core facts for rubric induction.

\textbf{Stage 3---Rubric induction.} Consolidated facts are used to induce a draft grading rubric under a system prompt that specifies explicit tier semantics---distinguishing criteria required for credit from criteria that are supplementary. Immediately after induction, a deterministic validation-and-repair step enforces formatting, structural, and rubric-schema constraints (e.g., that every tier is well-formed and grounded in the supplied facts) before the rubric proceeds to quality evaluation. This deterministic pass is distinct from, and runs prior to, the LLM-based structural repair described under Stage 4.

\textbf{Stage 4---Quality evaluation.} A quality-check stage, powered by GLM-5.1, scores the rubric on four dimensions---specificity, atomicity, tier correctness, and coverage---and recommends one of three actions:

\begin{itemize}
\item \textbf{Structural repair}, for defects fixable by reorganizing existing content. Repair is explicitly scoped: it may split, merge, re-tier, or reword existing criteria, but the system prompt's hard constraints prohibit introducing any new fact, number, or claim not already present in the rubric being repaired. Repair is retried up to a fixed maximum attempts, each attempt rechecked, with the best-scoring attempt kept and weaker or invalid attempts discarded.
\item \textbf{Targeted enrichment}, when coverage gaps cannot be closed by restructuring because they require new evidence. Rather than revising rubric wording, the system constructs a gap specification from the evaluator's feedback and re-examines the original model answer files to extract additional candidate facts addressing the missing concepts specifically. Newly extracted facts pass through the same agreement-based consolidation logic used in Stage 2, before being merged into the existing fact pool and the rubric re-induced.
\item \textbf{Stop}, when the rubric already meets the quality threshold.
\end{itemize}

\textbf{Hallucination safeguards during repair.} To reduce hallucination risk during structural repair specifically, the system performs numeric-payload-preservation checks: any repaired criterion introducing a numeric quantity, percentage, or monetary value not present anywhere in the rubric being repaired is flagged, that repair attempt is discarded, and the system either retries or reverts to the pre-repair rubric. This check compares the repaired rubric against the immediately preceding version of the rubric being repaired, not against the source filing directly---grounding in the filing is established earlier, at the fact-extraction stage, rather than re-verified at every repair step.

\textbf{Deduplication.} Following enrichment specifically, a deduplication pass clusters and canonicalizes near-duplicate criteria---which the enrichment step's merging of newly extracted facts into the existing pool can introduce---using the same fact-clustering logic as Stage 2, with each merged cluster inheriting the highest-priority tier among its members. The enriched-and-deduplicated rubric is then sent back to Stage 4 for re-evaluation. Rubrics routed to repair are instead rechecked directly, without passing through this deduplication step; rubrics that received a stop recommendation skip both repair/enrichment and deduplication entirely and proceed straight to expert review.

\subsection{Human Expert Review}

Rubrics that converge through the automated pipeline are not deployed directly. A human expert reviews each final rubric against the source filing before it is used for grading, providing a check on systematic biases the automated pipeline might share across its component models (for instance, an industry-wide tendency among extractor models to over- or under-weight certain disclosure types) that an evaluator built from the same model family might not catch on its own. This expert-review step is the final stage before a rubric is considered production-ready for benchmark grading.

\section{Results}
\label{sec:results}

\subsection{Overall Performance}
\label{sec:overall-perf}

Table~\ref{tab:results1} reports results for six tested models on the publicly released SpaceX question set ($n=70$). Scores are the mean per-question rubric score (0--1, reflecting partial credit against the expert-reviewed rubric), alongside the minimum and maximum per-question scores observed for each model. Cost-per-query figures are reported when available.

\begin{table}[h]
\centering
\caption{IPO Finance Agent results on the public SpaceX question set ($n=70$).}
\label{tab:results1}
\begin{tabular}{lccccc}
\toprule
Model & Mean score & Min & Max & Cost / query \\
\midrule
GLM-5.2          & 79.8\% & 44.9\% & 100\% & ---\textsuperscript{$\dagger$} \\
GLM-5.1          & 79.7\% & 24.8\% & 97.1\% & ---\textsuperscript{$\dagger$} \\
Qwen 3.7 Max     & 78.9\% & 17.8\% & 95.8\% & \$0.30 \\
MiMo-2.5 Pro     & 77.2\% & 47.3\% & 100\% & \$0.05 \\
Kimi K2.6        & 68.0\% & 0.0\%  & 100\% & \$0.21 \\
Nemotron 3 Ultra & 45.5\% & 0.0\%  & 95.8\% & \$0.14 \\
\bottomrule
\end{tabular}
\vskip 0.5ex
{\footnotesize \textsuperscript{$\dagger$}Sponsored API access via Z.AI; cost not applicable.}
\end{table}

Four models---GLM-5.2, GLM-5.1, Qwen 3.7 Max, and MiMo-2.5 Pro---cluster within three points of one another at the top of the leaderboard (77.2--79.8\%), suggesting a degree of capability convergence among current frontier and near-frontier models on this task once a sufficiently capable agentic harness is in place. GLM-5.2 leads the benchmark, though the evaluator (GLM-5.1) and the top-scoring model share the same model family; self-serving evaluation bias is a documented concern in LLM-as-a-judge systems \cite{zheng2023,liu2023geval,zhu2024judgelm}, and the GLM-5.2 result should be read with that caveat in mind. Kimi K2.6 trails this cluster by roughly ten points, and NVIDIA Nemotron 3 Ultra trails substantially further (45.5\%), with both Kimi K2.6 and Nemotron 3 Ultra recording a minimum per-question score of 0.0\%---indicating complete failure on at least one question, compared to a higher floor (17.8\% and above) for the top four models.

\subsection{Comparison to Finance Agent v2}

Table~\ref{tab:results2} places these results alongside the current Finance Agent v2 leaderboard, which evaluates periodic-filing (10-K/10-Q) tasks using a naive chunk-retrieval harness.

\begin{table}[h]
\centering
\caption{IPO Finance Agent vs.\ Finance Agent v2 leaderboards.}
\label{tab:results2}
\begin{tabular}{llcc}
\toprule
Benchmark & Model & Accuracy & Cost / query \\
\midrule
IPO Finance Agent & GLM-5.2            & 79.8\% & --- \\
IPO Finance Agent & Qwen 3.7 Max       & 78.9\% & \$0.30 \\
IPO Finance Agent & MiMo 2.5 Pro       & 77.2\% & \$0.05 \\
Finance Agent v2  & Gemini 3.5 Flash    & 57.9\% & \$2.51 \\
Finance Agent v2  & Claude Fable 5      & 56.3\% & \$8.06 \\
Finance Agent v2  & Claude Opus 4.8     & 53.9\% & \$4.22 \\
Finance Agent v2  & GPT-5.5             & 51.8\% & \$4.15 \\
Finance Agent v2  & MiniMax M3             & 48.3\% & \$0.32 \\
\bottomrule
\end{tabular}
\end{table}

GLM models, Qwen 3.7 Max and MiMo-2.5 Pro-based  IPO Finance Agents exceed every model on the current Finance Agent v2 leaderboard on accuracy, by a margin of at least 19 points over the FABv2 ceiling (Qwen 3.7 Max vs.\ Gemini 3.5 Flash). The gap is larger still on cost: MiMo-2.5 Pro exceeds FABv2's \emph{cheapest} competitive entry (MiniMax M3, 48.3\% at \$0.32) on accuracy while costing six times less per query.

This comparison should be read with a caveat, an ablation study would be required for an apple-to-apples comparison. In particular, while both benchmarks score on a partial-credit basis rather than strict exact-match, Finance Agent v2's Partial Credit metric is dealbreaker-gated---a response scores zero if it fails any load-bearing check, regardless of partial correctness elsewhere---whereas our per-question score is a continuous rubric-weighted average without an equivalent gating mechanism. The two metrics are not equivalent, and the gap reported above should be interpreted accordingly.

\subsection{Cost-Accuracy Tradeoff}

Cost-per-query data for four of the six tested models is reported in Table~\ref{tab:results1}, allowing a within-benchmark Pareto analysis. MiMo-2.5 Pro (77.2\% at \$0.05) strictly dominates both Kimi K2.6 (68.0\% at \$0.21) and Nemotron 3 Ultra (45.5\% at \$0.14)---it is simultaneously cheaper and more accurate than either. The efficient frontier among models with known cost therefore consists of just two points: MiMo-2.5 Pro at the low-cost end and Qwen 3.7 Max (78.9\% at \$0.30) at the high-accuracy end, with Kimi K2.6 and Nemotron 3 Ultra both Pareto-dominated. GLM-5.2 and GLM-5.1 cannot currently be placed on this frontier since their costs are unknown, having been evaluated via sponsored API access provided by Z.AI rather than standard metered pricing.

\subsection{Breakdown by Domain and Professional Workflow}

Table~\ref{tab:domain} reports mean per-question scores by the seven-domain taxonomy of Section~\ref{sec:dataset}; Table~\ref{tab:workflow} reports the corresponding breakdown by the six professional workflows of Section~\ref{sec:dataset}. Both decompose the same 70-question public set used in Table~\ref{tab:results1}, so per-cell sample sizes are necessarily small and should be read with that in mind---governance and control structures ($n=4$) and investment banking / equity capital markets ($n=5$) are the thinnest slices.

\begin{table}[h]
\centering
\footnotesize
\caption{Mean score by IPO-diligence domain ($n=70$).}
\label{tab:domain}
\begin{tabular}{lcccccc c}
\toprule
Domain & Qwen 3.7 Max & GLM-5.1 & GLM-5.2 & MiMo-2.5 Pro & Kimi K2.6 & Nemotron 3U & $n$ \\
\midrule
Segment economics \& operating performance & 77.9\% & 79.6\% & \textbf{82.7\%} & 80.9\% & 71.6\% & 62.4\% & 10 \\
KPI quality and monetization & 82.1\% & \textbf{82.2\%} & 81.4\% & 72.2\% & 73.9\% & 62.4\% & 12 \\
Governance and control structures & 78.9\% & 81.2\% & \textbf{85.3\%} & 72.3\% & 62.8\% & 49.5\% & 4 \\
Accounting and common-control recast mechanics & \textbf{79.9\%} & 76.5\% & 79.1\% & 75.6\% & 62.2\% & 41.3\% & 10 \\
Capital intensity and funding & \textbf{83.2\%} & 82.9\% & 83.0\% & 80.5\% & 74.2\% & 47.0\% & 12 \\
Execution and program risk & 82.8\% & \textbf{83.2\%} & 77.4\% & 82.7\% & 76.6\% & 26.9\% & 8 \\
Valuation and underwriting analysis & 70.5\% & 74.7\% & 74.0\% & \textbf{75.4\%} & 55.9\% & 30.1\% & 14 \\
\bottomrule
\end{tabular}
\end{table}

No single model leads all seven domains. GLM-5.2 tops three domains (segment economics, governance, and---by a wide margin---governance and control structures at 85.3\%); GLM-5.1 leads two (KPI quality and execution and program risk); Qwen 3.7 Max leads two (accounting and common-control recast mechanics, and capital intensity and funding); and MiMo-2.5 Pro leads one (valuation and underwriting analysis at 75.4\%). Valuation and underwriting analysis remains the most contested and lowest-scoring domain for most models: Qwen 3.7 Max, GLM-5.2, Kimi K2.6, and Nemotron 3 Ultra all post their weakest result there, while GLM-5.1 and MiMo-2.5 Pro do not. NVIDIA Nemotron 3 Ultra's gap to the top-scoring model is widest in execution and program risk (56 points) and valuation and underwriting analysis (45 points), and narrowest in KPI quality and monetization (20 points), its strongest relative domain.

\begin{table}[h]
\centering
\footnotesize
\caption{Mean score by professional workflow ($n=70$).}
\label{tab:workflow}
\begin{tabular}{lcccccc c}
\toprule
Workflow & Qwen 3.7 Max & GLM-5.1 & GLM-5.2 & MiMo-2.5 Pro & Kimi K2.6 & Nemotron 3U & $n$ \\
\midrule
Investment banking / ECM & \textbf{77.1\%} & 72.9\% & 76.7\% & 75.2\% & 56.2\% & 45.6\% & 5 \\
Public-market investing & 73.4\% & 76.9\% & 76.7\% & \textbf{78.1\%} & 66.3\% & 45.4\% & 19 \\
Venture capital and growth investing & 81.9\% & 82.4\% & \textbf{82.6\%} & 75.4\% & 76.9\% & 49.8\% & 12 \\
Credit analysis & 79.5\% & 79.5\% & \textbf{79.6\%} & 78.5\% & 67.5\% & 37.5\% & 12 \\
Securities counsel & 82.4\% & 82.2\% & \textbf{83.7\%} & 76.7\% & 66.8\% & 56.1\% & 6 \\
Accounting and transaction advisory & 82.2\% & \textbf{82.2\%} & 81.2\% & 77.3\% & 67.9\% & 44.5\% & 16 \\
\bottomrule
\end{tabular}
\end{table}

The workflow breakdown distributes leadership more broadly. GLM-5.2 leads three of six workflows (venture capital and growth investing, credit analysis, and securities counsel), while MiMo-2.5 Pro leads public-market investing (78.1\%), Qwen 3.7 Max leads investment banking / ECM (77.1\%), and GLM-5.1 ties for first in accounting and transaction advisory (82.2\%, equal to Qwen 3.7 Max). Kimi K2.6 and Nemotron 3 Ultra place fifth and sixth, respectively, on every workflow without exception.

\section{Discussion}
\label{sec:discussion}

\subsection{Harness Quality Versus Question Difficulty}
\label{sec:harness-quality}

A central claim motivating this paper is that IPO due-diligence reasoning over governance, common-control recast accounting, and secondary-market valuation arguably demands at least as much multi-step synthesis as Finance Agent v2's hardest categories (Financial Modeling, Precedents), which top out at roughly 23\% accuracy on that benchmark, yet IPO Finance Agent's tested models substantially exceed FABv2's accuracy ceiling (Sections~\ref{sec:results}.2--\ref{sec:results}.3). We find this pattern consistent with a harness-quality explanation rather than an easier-question explanation: four independent model families (Qwen, GLM, MiMo, and Kimi at the upper end) cluster within a relatively narrow band near 68--80\%, which is more consistent with a harness that reliably surfaces the right evidence---such that multiple sufficiently capable models can act on it correctly---than with a benchmark whose difficulty is dominated by question design alone. This interpretation is broadly consistent both with findings from the agent-evaluation literature, where environment design, retrieval quality, tool availability, and evaluation methodology often explain a substantial fraction of observed performance differences between systems built on otherwise comparable foundation models \cite{liang2022helm,liu2023agentbench}, and with evidence that retrieval quality specifically tends to dominate downstream performance on long-document question-answering tasks \cite{lewis2020rag,anthropic_contextual_retrieval}.

However, this comparison is not a direct or controlled one. Finance Agent v2 and IPO Finance Agent differ in task domain, question taxonomy, document type, and grading metric. We have not yet run an ablation that isolates the contribution of contextual retrieval by, for instance, evaluating the same models on the same question set with and without it. This is left to future work.

\section{Conclusion}
\label{sec:conclusion}

We introduced IPO Finance Agent, extending the Finance Agent benchmark from periodic-reporting analysis to IPO due-diligence reasoning over S-1 filings, using the SpaceX IPO filing as a public case study. Our contribution is threefold: a seven-domain, workflow-labeled question taxonomy designed around the specific reasoning demands of IPO filings rather than periodic filings; a contextual-retrieval harness that addresses the retrieval failures exhibited by naive chunking on long, cross-referential filings; and an automated, evaluator-optimizer rubric generation pipeline that produces grading criteria with substantially reduced human intervention. On the 70-question SpaceX set, the best-performing tested models exceed the current Finance Agent v2 leaderboard ceiling by a wide margin on both accuracy and cost. We view this pattern as consistent with the hypothesis that harness design and retrieval architecture materially affect performance on long registration statements, while noting that the contribution of individual components has not yet been isolated through controlled ablation.

Several directions for future work follow: multi-run variance reporting to align with Finance Agent v2's measurement standard; tool-call analysis to test whether the search-versus-calculation allocation pattern observed in Finance Agent v2 holds in this setting; and a more rigorous ablation isolating the contribution of contextual retrieval from other harness differences. This pattern situates IPO Finance Agent within a wider transition from static benchmark evaluation toward workflow-grounded agent evaluation, where success depends jointly on retrieval, tool use, evidence synthesis, and domain-specific reasoning rather than isolated question answering alone \cite{liu2023agentbench,yao2024taubench,jimenez2024swebench}. More broadly, we position S-1 analysis---and IPO due diligence as a professional workflow---as a high-value, currently underbenchmarked testbed for evaluating whether LLMs can act as workflow-aware financial research agents rather than narrow question-answering systems, and we hope the rubric generation methodology introduced here proves useful for other benchmarks.

\section*{Acknowledgments}

We thank Cunxiang Wang from the Zhipu AI evaluation team for sponsoring API access to GLM-5.1 and GLM-5.2.

\bibliographystyle{plain}

\end{document}